# REAL TIME FACIAL EXPRESSION RECOGNITION USING A NOVEL METHOD

Saumil Srivastava[1]

[1]Department of Computer Science, Jaypee Institute of Information Technology, Noida, India
`saumil.srivastava@jiitu.org`

## ABSTRACT

*This paper discusses a novel method for Facial Expression Recognition System which performs facial expression analysis in a near real time from a live web cam feed. Primary objectives were to get results in a near real time with light invariant, person independent and pose invariant way. The system is composed of two different entities trainer and evaluator. Each frame of video feed is passed through a series of steps including haar classifiers, skin detection, feature extraction, feature points tracking, creating a learned Support Vector Machine model to classify emotions to achieve a tradeoff between accuracy and result rate. A processing time of 100-120 ms per 10 frames was achieved with accuracy of around 60%. We measure our accuracy in terms of variety of interaction and classification scenarios. We conclude by discussing relevance of our work to human computer interaction and exploring further measures that can be taken.*

## 1. INTRODUCTION

Major component of human communication are facial expressions which constitute around 55 percent of total communicated message [1]. We use facial expressions not only to express our emotions, but also to provide important communicative cues during social interaction, such as our level of interest, our desire to take a speaking turn and continuous feedback signaling understanding of the information conveyed.

There has been a global rush for facial expression recognition over the last few years. A number of methods have been proposed but no single method which is both efficient in terms of memory and time complexity has yet been found.

In 1978, Paul Ekman and Wallace V. Friesen published the Facial Action Coding System (FACS)[14], which, 30 years later, is still the most widely used method available. Through observational and electromyography study of facial behavior, they determined how the contraction of each facial muscle, both singly and in unison with other muscles, changes the appearance of the face. It talks about six basic emotions (anger, disgust, fear, joy, sorrow, surprise). FACS codes expressions using a combination of 44 facial movements called as action units (AU"s).While lot of work on FACS has been done and also FACS being an efficient, objective method to describe facial expressions, it is not without its drawbacks. Coding a subject's video is a time- and labor-intensive process that must be performed frame by frame. A trained, certified FACS coder takes on average 2 hours to code 2 minutes of video. In situations where real-time feedback is desired and necessary, manual FACS coding is not a viable option.

This gives rise to our problem statement i.e. to be able to infer emotions in real time using a live video feed with the following challenges

- To be able to provide results in real time
- To be able to make a person independent model
- To be efficient in classifying the emotions into one of the following anger, smile, surprise, neutral.
- Light Invariant Application
- Pose Invariant

Pantic and Rothkrantz[2] suggested three basic problems for expression analysis, the three steps that can be identified as three separate problems in the whole process i.e. face detection in a facial image or image sequence ,facial expression data extraction and facial expression classification. Face detection methods that have been around mostly assumes frontal face view in image or image sequence being analyzed. Viola & Jones [3] provides competitive face detection in real time, uses the adaboost algorithm to exhaustively pass a search window over the image at various scales for rapid detections. Essa & Pentland[4] perform spatial and temporal filtering together with thresholding to identify the motion blobs from image sequence. To detect face then Eigen face method [5] is used .The PersonSpotter system [6] tracks the bounding box of the head in video using spatio-temporal filtering and disparity in pixels ,thereby selecting the ROI's and then passing through skin detector and convex region detector to check for face.

Second step involves face data extraction, Littleworth et al. [7] used a bank of 40 gabor filters at different scale and orientations to apply convolution on the image and then complex value response was recorded as in [8]. Essa & Pentland [4] use the same face extraction image to find facial feature points using a set of sample images via FFT and local energy computation. Cohn et al. [9] firstly normalize the feature points in first frame of image sequence and then use optical flow to track them .The displacement vectors for each landmark between the initial and peak frame represent the extracted information. Other methods include AAM, T.F.Cootes [10] et al. uses a statistical model of shape and grey-level appearance of object of interest which can generalize to almost any valid example .Match is obtained in a few iterations by changing parameters in accordance to residual of trained and current image.

Third step involves classification of emotions based on a training model constructed on the properties related to feature points for this purpose we have created our learning model using support vector machines[11]

## 2. IMPLEMENTATION

The overall system has been developed using C/C++ with support of OpenCV, libSVM[11] libraries. The overall process can be shown as follows

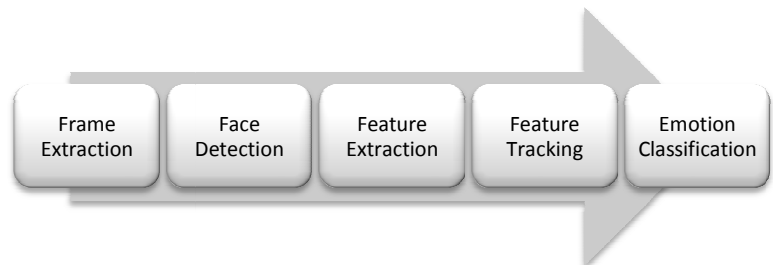

Figure 1. The overall process for facial expression recognition

## 2.1. Detection of Haar Like Features (Viola & Jones [3])

Viola and Jones in 2001 published their breakthrough work which allowed appearance based methods to run in real time, while keeping the same or improved accuracy.

**Rectangle Features**

The sums of the pixels which lie within the white rectangles are subtracted from the sum of pixels in the grey rectangles.

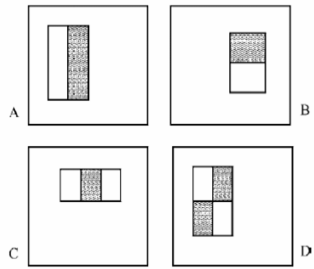

Figure 2. Examples of Haar-like feature sets.

Rectangle Features can be computed very rapidly using an intermediate representation called integral image. The value of the integral image at point (x,y) is the sum of all the pixels above and to the left.

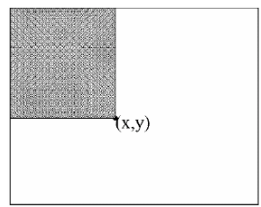

Figure 3. The value of the integral image at point (x,y) is the sum of all the pixels above and to the left.

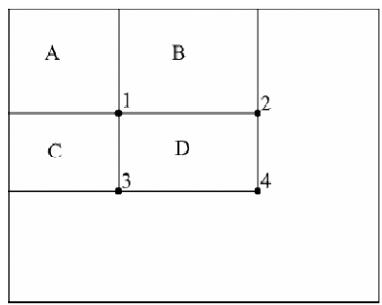

Figure 4. The sum of pixels within D can be computed as 4+1-(2+3)

Scan image at multiple positions and scales as in previous approaches. Apply Adaboost strong classifier (which is based on Rectangle Features) to decide whether the search window contains a face or not.

Adaboost Strong Classifier: linear combination of weak classifiers. H(x) = 1 if window x has a face and 0 otherwise

$$H(x) = \begin{cases} 1 \text{ if } \sum_{t=1}^{T} \alpha_t h_t(x) \geq \phi \\ 0 \text{ otherwise} \end{cases}$$
→ Threshold
→ Weak Classifier
→ Weights

$$h_j(x) = \begin{cases} 1 \text{ if } p_j f_j(x) < p_j \theta_j \\ 0 \text{ otherwise} \end{cases}$$
→ Threshold
→ Sign
→ Rectangle Feature

Each week classifier corresponds to a single Rectangle Feature.

Adaboost ensembles many weak classifiers into one single strong classifier

1. Initialize sample weights

2. For each cycle:

Find a classifier/rectangle feature that performs well on the weighted samples.

Increase weights of misclassified examples.

3. Return a weighted combination of classifiers

Attentional Cascade

1. We start with simple classifiers which reject many of the negative subwindows

2. While detecting almost all positive sub-windows

3. Positive results from the first classifier triggers the evaluation of a second (more complex) classifier, and so on

4. A negative outcome at any point leads to the immediate rejection of the subwindow

6. The attentional cascade is based on the assumption that within an image, most sub-images are non-face instances.

## 2.2. Face Detection

Each frame is processed firstly through Haar classifiers [3] trained for profile faces. To further improve frame rate and compensate for pose variation. We propose to use interleaved Haar Classifiers. Interleaving is done between front and profile classifiers.

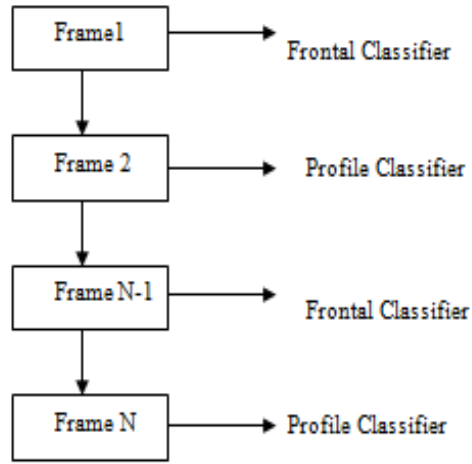

Figure 5. Interleaved Classifiers

Second to reduce false results produce, output of last step is passed through skin detection algorithm [13].This step makes sure that the region selected is a face. If both profile and frontal faces are not detected and if face has been initialized then Optical Flow [9] is applied to track those feature points which is discussed below.

## 2.3. Facial Feature Point Extraction

After face has been detected, we narrow our work zone to detected region thereby reducing the search area and improving frame rate. Now face is divided geometrically to find feature points and further confine the search area.

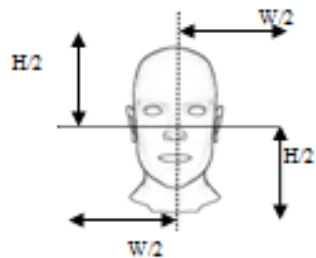

Figure 6. Geometrically Divided Face

Then classifiers for eyes, nose, and mouth are used in the narrowed output region. After that sobel derivative is applied on the region, The function calculates the image derivative by convolving the image with the appropriate kernel

$$\mathtt{dst}(x,y) = \frac{d^{xorder+yorder}\mathtt{src}}{dx^{xorder} \cdot dy^{yorder}}$$

The Sobel operators combine Gaussian smoothing and differentiation so the result is more or less resistant to the noise. Most often, the function is called with ( xorder = 1, yorder = 0, apertureSize = 3) or ( xorder = 0, yorder = 1, apertureSize = 3) to calculate the first x- or y- image derivative.
After that corners are detected using Shi-Tomasi algorithm[12], The function finds the corners with big eigen values in the image. The function first calculates the minimal eigen value for every source image pixel and stores them. Then it performs non-maxima suppression (only the local maxima in neighborhood are retained). The next step rejects the corners with the minimal eigen value less than quality level of maximum eigen value . Finally, the function ensures that the distance between any two corners is not smaller than minDistance. The weaker corners (with a smaller min eigen value) that are too close to the stronger corners are rejected , thus 21 feature points are selected (Fig 7.b). If face is not detected by classifier then based on previous 21 feature vector using Optical Flow method[9] (Fig 7.c).

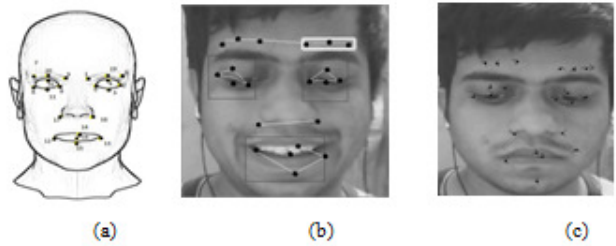

Figure 7. (a) 21 feature points on face (b) When face detected(c) When face not detected

Optical Flow states that given a set of points in an image, find those same points in another image. or, given point [ux, uy]T in image I 1 find the point [ux + δx, uy + δy]T in image I 2 that minimizes ε:

$$\varepsilon(\delta_x, \delta_y) = \sum_{x=u_x-w_x}^{u_x+w_x} \sum_{y=u_y-w_y}^{u_y+w_y} \left(I_1(x,y) - I_2(x+\delta_x, y+\delta_y)\right)$$

(the Σ/w's are needed due to the aperture problem)

Median of 21 points over the next ten frames is taken so as to account for any error in calculation that may be occurring like it's not always necessary that we detect all 21 points in which frame, to reduce such error we take median over the 10 frame intervals hence reducing error.

The overall process can be summarized as below

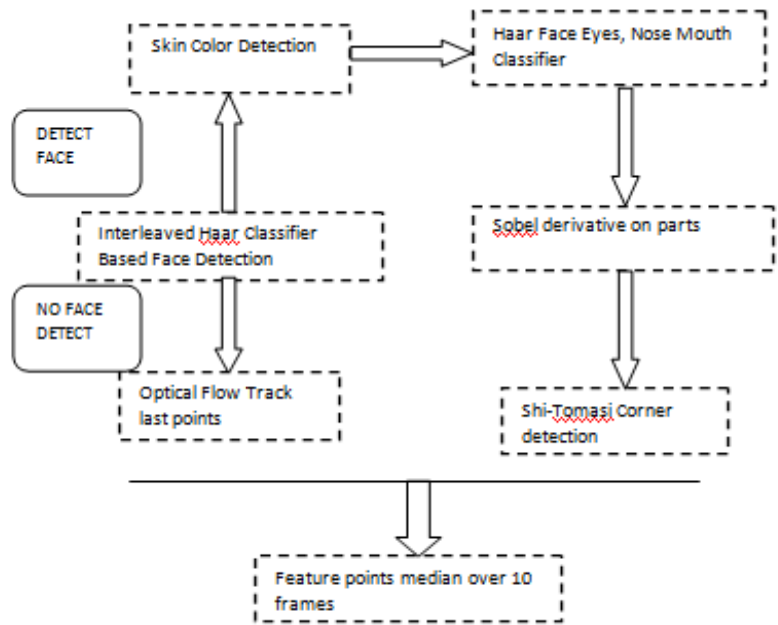

Figure 8. Over all Feature Extraction Diagram

## 3. SVM CLASSIFICATION

Displacements of 21 feature points is used as an input to SVM classifier first a model is created with help of libSVM[11] by training and then Evaluator is used to classify emotions based on the learned model created.

Following steps for using libSVM were used

- Transform data to the format of an SVM package
- Conduct simple scaling on the data
- Consider the RBF kernel $K(\mathbf{x}, \mathbf{y}) = e^{-\gamma \|\mathbf{x}-\mathbf{y}\|^2}$
- Use cross validation to find the best parameter C and gamma
- Use the best parameter C and gamma to train the whole training set
- Test

This kernel nonlinearly maps samples into a higher dimensional space so it, unlike the linear kernel, can handle the case when the relation between class labels and attributes is nonlinear. Furthermore, the linear kernel is a special case of RBF.

The second reason is the number of hyperparameters which influences the complexity of model selection. The polynomial kernel has more hyperparameters than the RBF kernel.

Most computational overhead resides in the training phase. However, due to the fact that the training set is interactively created by the user and hence limited in magnitude and that the individual training examples are of constant and small size, overhead is low for typical training runs.

Training on different individuals lead to model being more of person independent nature. For training purposes 5 subjects 4 expressions (Neutral, Smile, Angry, and Excited) were used.

## 4. RESULTS

The confusion matrix based on various interaction clips is shown in table below

|  | Neutral | Smile | Angry | Excited | Over all |
|---|---|---|---|---|---|
| Neutral | 15 | 3 | 12 | 0 | 50% |
| Smile | 5 | 18 | 5 | 2 | 60% |
| Angry | 10 | 5 | 13 | 2 | 43.3% |
| Excited | 0 | 4 | 0 | 26 | 86.67% |
|  |  |  |  | Total | 59.91% |

## 4. CONCLUSION AND FUTURE WORK

Average frame processing time 120-150ms per 10 frames. It was observed that result was directly proportional to intensity of training provided. It was also observed that result was robust to pose variations; light intensity changes and is person independent model. Also the result was robust to background changes.

Further work : To minimize the effect of camera distance from user ,also to make database more extensive so as to improve results, link with Paul Ekman's FACS Action Units [14 ] to increase range of emotions to be tracked and also to be more accurate.

Inclusion of emotions in human computer interface is an emerging field. It provides us with many new opportunities. Development of computationally effective and robust solutions will lead to increased importance of user in the process and set stage for revolutionary interactivity.

### ACKNOWLEDGEMENTS

Dr. Krishna Asawa, Associate Professor, Jaypee Institute of Information Technology, Noida mentored me throughout the project and without her project would not have been possible. Also resources made available to me through Affective Computing Lab Jaypee Institute of Information Technology were valuable in the course of the project.

### REFERENCES

[1]    A. Mehrabian. Communication without words. Psychology Today, 2(4):53–56,(1968)

[2]    M. Pantic and L. J. M. Rothkrantz. Automatic analysis of facial expressions: The state of the art . IEEE Transactions on Pattern Analysis and Machine Intelligence, 22(12):1424–1445, (2000).

[3]    P. Viola and M. Jones, "Robust Real-Time Face Detection," Int'l J. Computer Vision, vol. 57, no. 2, pp. 137-154, May 2004.

[4]  I. Essa and A. Pentland. Coding, analysis,interpretation and recognition of facial expressions. IEEE Transactions on Pattern Analysis and Machine Intelligence, 19(7):757–763, (1997).

[5]  M. Turk and A. Pentland. Eigenfaces for recognition. Journal of Cognitive Neuroscience, 3(1):71–86, (1991).

[6]  J. Steffens, E. Elagin, and H. Neven. Personspotter—fast and robust system for human detection, tracking, and recognition. In ProceedingsInternational Conference on Automatic Face and Gesture Recognition, pages 516–521, (1998).

[7]  G. Littlewort, I. Fasel, M. Stewart Bartlett, and J. R. Movellan. Fully automatic coding of basic expressions from video. Technical Report 2002.03, UCSD INC MPLab, (2002).

[8]  M. Lades, J. C. Vorbr¨uggen, J. Buhmann, J. Lange, C. von der Malsburg, R. P. W¨urtz, and W. Konen. Distortion invariant object recognition in the dynamic link architecture. IEEE Transactions on Computers, 42:300–311, (1993).

[9]  J. F. Cohn, A. J. Zlochower, J. J. Lien, and T. Kanade. Feature-point tracking by optical flow discriminates subtle differences in facial expression. In Proceedings International Conference on Automatic Face and Gesture Recognition, pages 396–401 (1998).

[10] T.F.Cootes, G.J. Edwards and C.J.Taylor. "Active Appearance Models", in Proc. European Conference on Computer Vision 1998 (H.Burkhardt & B. Neumann Ed.s). Vol. 2, pp. 484-498, Springer (1998).

[11] C.-C. Chang and C.-J. Lin. LIBSVM: a library for support vector machines,(2001). Software available at http://www.csie.ntu.edu.tw/~cjlin/libsvm.

[12] Jianbo Shi and Tomasi, C."Good Features To Track" inComputer Vision and Pattern Recognition, 1994. Proceedings CVPR '94 (1994)

[13] Dadgostar Farhad, and Sarrafzadeh Abdolhossein,An adaptive real-time skin detector based on Hue thresholding: A comparison on two motion tracking methods,2006 – Elsevier

[14] Andrew Ryan ,Jeffery F. Cohn, Simon Lucey, Jason Saragih, Patrick Lucey, Fernando De la Torre and Adam Rossi. "Automated Facial Expression Recognition System" Security Technology, 2009. 43rd Annual 2009 International Carnahan Conference (2009).

[15] Open CV http://opencv.willowgarage.com/wiki/

**Authors**

Student at Jaypee Institute of Information Technology. Currently pursuing MBA in Finance and International Business from Jaypee Business School as a part of Dual Degree Course. Computer Science Engineer with interests in Affective Computing, Human Computer Interaction, Optimization Techniques, Computer Vision.

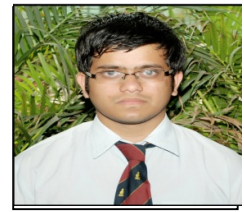